\documentclass[final]{cvpr}

\usepackage{times}
\usepackage{epsfig}
\usepackage{graphicx}
\usepackage{amsmath}
\usepackage{amssymb}
\DeclareMathOperator*{\argmax}{arg\,max}
\newtheorem{theorem}{Theorem}[section]
\newtheorem{lemma}[theorem]{Lemma}
\usepackage[pagebackref=true,breaklinks=true,colorlinks,bookmarks=false]{hyperref}

\def\cvprPaperID{9} % *** Enter the CVPR Paper ID here
\def\confYear{CVPR 2021}
\begin{document}

%%%%%%%%% TITLE
% \title{Instance-wise Causal Feature Selection}

\title{Instance-wise Causal Feature Selection for Model Interpretation}

\author{Pranoy Panda \\
IIT Hyderabad\\
{\tt\small cs20mtech12002@iith.ac.in}
\and
Sai Srinivas Kancheti\\
IIT Hyderabad\\
{\tt\small cs21resch01004@iith.ac.in}
\and
Vineeth N Balasubramanian\\
IIT Hyderabad\\
{\tt\small vineethnb@iith.ac.in}
}
% {\small\{cs20mtech12002, cs21resch01004, vineethnb\}@iith.ac.in}

% \and

\maketitle

%%%%%%%%% ABSTRACT
\begin{abstract}
   We formulate a causal extension to the recently introduced paradigm of \textit{instance-wise feature selection} to explain black-box visual classifiers. Our method selects a subset of input features that has the greatest causal effect on the model's output. We quantify the causal influence of a subset of features by the Relative Entropy Distance measure. Under certain assumptions this is equivalent to the conditional mutual information between the selected subset and the output variable. The resulting causal selections are sparser and cover salient objects in the scene. We show the efficacy of our approach on multiple vision datasets by measuring the post-hoc accuracy and Average Causal Effect of selected features on the model's output.   
\end{abstract}

%%%%%%%%% BODY TEXT
\vspace{-10pt}
\section{Introduction}

Explaining the predictions of black-box classifiers is important for their integration in real world applications. There have been many efforts to understand the predictions of visual systems, by generating saliency maps that quantify the importance of each pixel to the model's output. However, such methods usually require gradient information and often suffer from insensitivity to the model and the data~\cite{Adebayo2018SanityCF}. Researchers in the recent past have taken a different perspective by proposing the task of \textit{instance-wise feature selection} for explaining classifiers in general, and visual classifiers in particular. Here, the aim is to select a subset of pixels or superpixels to explain a black-box model's output.

%SotA
L2X\cite{chen2018learning} is the first work in this space wherein it selects a fixed number of features which maximize the mutual information w.r.t. the output variable. Its successor, INVASE\cite{yoon2018invase} removes the constraint of having to fix the number of features to be selected. But, both INVASE and L2X have optimization functions which try to maximize mutual information in some form or the other. For an explanation to be correct, the selected features should be causally consistent with the model being explained. Therefore, a good instance-wise feature selection method should capture the most causal features in an instance. We hypothesize that the most sparse and class discriminative features are indeed the most causal features, and they form good visual explanations. However, existing methods(L2X and INVASE) select features that may not capture causal influence, since mutual information does not always capture causal strength\cite{chang2020invariant}. 

In this work we take a step towards unifying causality and instance-wise feature selection to select causally important features for explaining a black box model's output. First, in order to measure causal influence of input features w.r.t the output, we choose a causal metric which satisfies properties relevant for our task and subsequently we simplify this metric(under certain assumptions) to conditional mutual information. Secondly, we derive an objective function for training our explainer using continuous subset sampling. We evaluate our explainer on 3 vision datasets and compare it with with 3 popular baseline explainability methods. For the purposes of quantitative comparison, we use two metrics, post-hoc accuracy\cite{chen2018learning} and a variant of average causal effect(ACE) which we introduce in our work. Our results show performance improvements over the baselines, especially in terms of the ACE values, which verifies our claim of selecting causal features.  
% We instead propose to explain visual classifiers via \textcolor{blue}{\textit{instance-wise}} \textit{feature selection}, where a subset of pixels or superpixels is chosen to explain the model's output. A limitation of existing methods \textcolor{blue}{in this space(L2X[ref] and INVASE[ref])} is that the selected features may not be causally consistent with the model to be explained. 

% In order to select features which have the greatest causal strength w.r.t the model output, we train a \textit{selector} that learns to choose an input subset which maximizes the Relative Entropy Distance.

% \textcolor{red}{Write about why causal subset selection is needed}  

\begin{figure*}[!t]
\vspace{-50pt}
\begin{center} 
\includegraphics[scale=0.7]{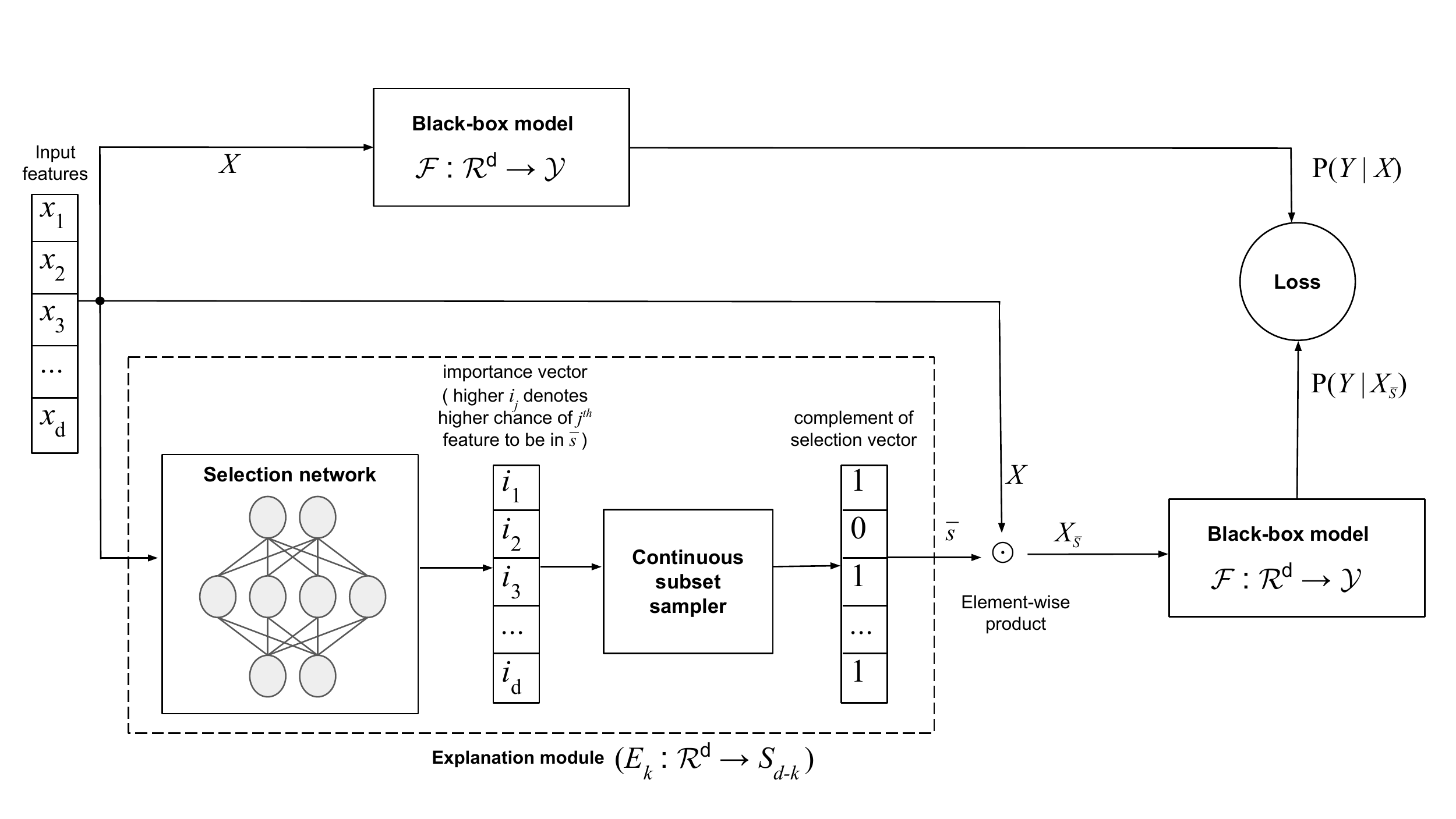}
\caption{\textbf{Block diagram of our method.} Samples of the training data are fed to the explanation module to get a $(d-k)$-hot random vector as output($k$ is the size of the subset that we want to select for explanation). Subset selection is done by generating a importance vector from the input features(via a neural network) and then using it along with Gumbel-softmax trick. Then, to compute the objective function we pass $X_{\overline{s}}$ and $X$  through the black-box model to evaluate the conditional probabilities. Finally, we backprop through the loss to update the selector network’s weights.}
\label{figbd}
\end{center}
\vspace{-5pt}
\end{figure*}

\subsection{Problem Formulation}
% Our goal is to explain a black-box classifier $\mathcal{F}: \mathcal{R}^{d}\rightarrow \mathcal{Y}$ by learning a selector network $S_k: \mathcal{R}^{d}\rightarrow E_k$ where $E_k=\{e| e\in\{0,1\}^m, \lvert e\rvert=k\}$. The selector network $S_k$ chooses a subset of fixed size $k$ for each input that best explains the predictions made by $\mathcal{F}$.  We choose RED to quantify the causal strength of a subset

% Let $\mathcal{X}= \mathcal{X}_1 \times \mathcal{X}_2 \times...\mathcal{X}_d$ be a $d$-dimensional feature space and $\mathcal{Y}={1,...,c}$ be a $c$-dimensional discrete label space. Let $X \in \mathcal{X}$, be an instance/data point and $Y \in \mathcal{Y}$ be its corresponding label. We also define a selection vector $s\in\{0,1\}^d$. If $s_i=1$, that means the $i^{th}$ feature is selected, else not.

% Our goal is to explain a black-box classifier $\mathcal{F}: \mathcal{R}^{d}\rightarrow \mathcal{Y}$ by learning a explainer $E_k: \mathcal{R}^{d}\rightarrow S_k$ where $S_k=\{e| e\in\{0,1\}^d, \lvert e\rvert=k\}$. The explainer $E_k$ chooses a subset of features(size of subset is fixed as $k$) for each input such that its complement best explains the predictions made by $\mathcal{F}$. The explanation is the complement of the selected subset(The reason for this will become clear in the Methodology section)
Our goal is to explain a black-box classifier $\mathcal{F}: \mathcal{R}^{d}\rightarrow \mathcal{Y}$ by learning a explainer $E_k: \mathcal{R}^{d}\rightarrow S_{d-k}$ where $S_{d-k}=\{e| e\in\{0,1\}^d, \lvert e\rvert=d-k\}$. The explainer $E_k$ determines a subset of features(size of subset is fixed as $k$) for each input which best explains the predictions made by $\mathcal{F}$. Please note that in practice our explainer chooses the complement of the explanation hence we end up selecting a $(d-k)$-hot vector. The actual explanation is determined by complementing this $(d-k)$-hot vector.

We intend to find the subset(of cardinality $k$) which has the maximum causal strength. Since there is no gold standard for measuring causal influence between random variables, we choose a metric which has some good properties. 

\begin{figure}[h]
\centering
\includegraphics[scale=0.25]{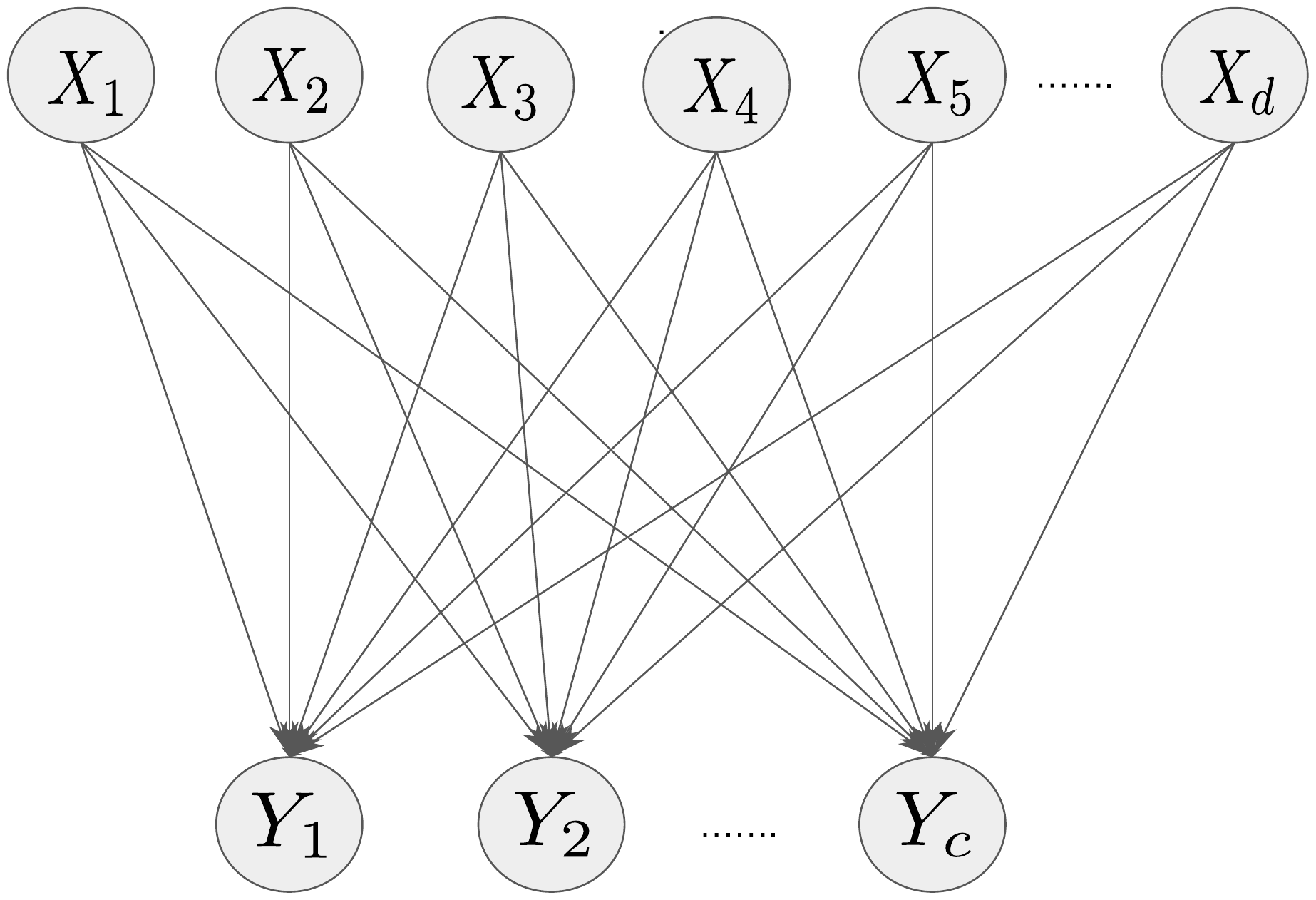}
\caption{Black-box model as a SCM}
\label{fig1}
\vspace{-10pt}
\end{figure}

\textbf{Causal Model:} Assuming that the underlying architecture of the black-box model is a directed acyclic graph(DAG), it can be shown that such models can be interpreted as structural causal models~\cite{chattopadhyay2019neural}. This SCM(Figure \ref{fig1}) simply has directed edges from input layer to the output layer representing the fact that the output is only a function of the inputs. We explain explicitly what $X_i$ and $Y_j$ mean in our context as we go along.
% To simplify our objective function we make a strong assumption that input features are independent.  

\textbf{Causal metric:} We now quantify the causal influence between two random variables in the causal graph(Figure \ref{fig1}) by the Relative Entropy Distance(RED)\cite{janzing2013quantifying} metric. RED is an information theoretic causal strength measure, which is based on performing interventions on the causal graph and exposing the target of the cut/intervened edges with product of marginal distributions.

We choose RED as the metric for causal strength because it satisfies the following useful properties:
\begin{itemize}
    % \item It can capture functional dependencies thus encapsulating the complex and nonlinear functions between input and output variables which are present in complicated models such as neural networks) (from \cite{o2020generative})
    \item It can capture the non-linear, complex relationship between input and output variables in a black-box model \cite{o2020generative}. Such complex relationships are common in models such as neural networks.
    \item The causal strength of an edge $X_i\rightarrow Y_j$ depends on $X_i$ and the other parents of $Y_j$ and on the joint distribution of parents of $Y_j$. This locality property ensures that we do not have to take into account the causes of $X_i$ when computing causal influence of $X_i$ on $Y_j$, unless the causes are also immediate causes of $Y_j$\cite{janzing2013quantifying}.
\end{itemize}

As we are operating in the vision domain, instead of reasoning at the level of subset of pixels, we reason at the level of superpixels/patches in an image. These patches are disjoint i.e. there is no overlap between the patches. Now, for simplifying the RED metric for the purposes of experimenting in real-world setting, we assume local influence-pixels only depend on other pixels within a patch. That is the reason why we have no edges between $X_{i}$'s in the SCM(Figure \ref{fig1}). Also, in the given SCM $X_i$ refers to the $i^{th}$ patch in the image and $Y_j$ refers to $j^{th}$ output node in the model. Under this setting we state lemma \ref{lemma1} due to \cite{janzing2013quantifying}.
\begin{lemma}
The causal strength of a subset of links in $s$ going from input $X$ to the response variable of the model $Y$, denoted as $CS_s$, is given as follows:
\begin{equation}
    CS_{s} =I( X_{s} ;Y|X_{\overline{s}})
    \label{eq1}
    \vspace{-8pt}
\end{equation}
\label{lemma1}
Here, $X_s$ denotes the features in set $s$, $X_{\overline{s}}$ denotes the features not in set $s$, and $X=X_{s}\cup X_{\overline{s}}$.
\vspace{-0.2cm}
\end{lemma}

% \noindent (Proof of lemma \ref{lemma1} is in the Appendix)

Now, we simplify the equation \ref{eq1} to formulate an optimization problem.

\textbf{Objective Function:} From lemma \ref{lemma1} we know that the causal strength of a set of input features $X_{s}$, denoted as $CS_s$, w.r.t. to the output $Y$ is given by the conditional mutual information. Now, we proceed to derive an optimization objective to find the subset with maximum causal strength.
\begin{gather*}
CS_{s} =I\left( X_{s} ;Y|X_{\overline{s}}\right) 
\end{gather*}
The above equation can be expressed explicitly in information theoretic terms as follows:
\begin{equation*}
CS_{s} =-H( Y|X) +H( Y|X_{\overline{s}}) 
\end{equation*}
In order to maximize the above objective function, we need to focus only on $H(Y|X_{\overline{s}})$ as the other term is independent of set $s$. Now, if we further simplify the remaining term by expanding it and viewing it as an expectation over variables $Y$ and $X_{\overline{s}}$, we get the following equation: 
% \begin{equation*}
%     CS_{s} =I\left( X_{s} ;Y|X_{\overline{s}}\right)   
% \end{equation*}
% \begin{equation*}
%     CS_{s} =-H( Y|X) +H( Y|X_{\overline{s}})
% \end{equation*}
% In order to solve the above objective function, we need to focus only on $H(Y|X_{\overline{s}})$ as the other term is independent of set $s$. Now, if we further simplify the remaining term by expanding it and viewing it as an expectation over variables $Y$ and $X_{\overline{s}}$, we get the following equation: 
\begin{align}
& \max_{s} CS_{s} \equiv \min_{s} E_{Y,X_{\overline{s}}}[\log( P( Y|X_{\overline{s}})] 
%\min_{s} E_{X} E_{\overline{s} |X} E_{Y|X_{\overline{s}}}\ \left[\log( p( Y|X_{\overline{s}})\right] 
% \boxed{ \begin{array}{{>{\displaystyle}l}}
% \max_{s} CS_{s} =\min_{s} E_{X_{\overline{s}}} E_{Y|X_{\overline{s}}}[\log( p( Y|X_{\overline{s}})]\\
% \ \ \ \ \ \ \ \ \ \ \ \ \ \ \ \ \ \ \ \ \ \ \ \ \ =\min_{s} E_{X} E_{\overline{s} |X} E_{Y|X_{\overline{s}}}[\log( p( Y|X_{\overline{s}})]\\
% \ \ \ \ \ \ \ \ \ \ \  =\min_{s} E_{Y,X_{\overline{s}}}[\log( p( Y|X_{\overline{s}})]
% \end{array}}\\
\label{eq2}
\end{align}
$\ P( \ \overline{s} |X\ )$, which is implicit in the above equation, is the selector's output distribution. 
\section{Methodology}
There are two main issues with maximizing the causal strength in equation \ref{eq2}. First, approximating the conditional distribution $P(Y|X_{\overline{s}})$, and second, dealing with subset sampling. We address these issues below.
\vspace{3pt}

\textbf{Approximating} $P(Y|X_{\overline{s}}):$ We simply use the output of the black-box model $\mathcal{F}$ when $X_{\overline{s}}$ is given as input, to estimate $P(Y|X_{\overline{s}})$. $X_{\overline{s}}$ is represented as follows: if $s_i=0$, ${X_{\overline{s}}}_i=X_i$, else ${X_{\overline{s}}}_i=0$. \cite{schwab2019cxplain} have used a similar approximation in their works.
\vspace{3pt}

\textbf{Continuous subset sampling:} Our objective function(\ref{eq2}) requires sampling of subsets which is a non-differentiable operation. Similar to \cite{chen2018learning} we use the Gumbel Softmax trick \cite{jang2016categorical, maddison2016concrete} for continuous subset sampling. 

The goal of this procedure is to sample a subset $s$ consisting of $k$ distinct features out of the $d$ input dimensions. Sampling set $s$ is similar to sampling a $k$-hot random vector, where the length of this random vector is $d$. Before we perform sampling, we define a function $g$ which maps each input feature $X_i$ to an importance score which indicates its probability of being part of $\overline{s}$. 
% In other words, $g(X)$ defines a categorical distribution from which we wish to sample from. 
We learn this function $g$ via a neural network parameterized by $\theta$. Then, we use the gumbel-softmax continuous subset sampling for sampling from the importance vector $g(X)$. The result of this sampling is a random variable $Z$ which is a function of the neural net parameters $\theta$ and Gumbel random variables $\zeta$. This effectively means that we can estimate $\overline{s}$ by $Z(\theta,\zeta)$.
\vspace{3pt}

\textbf{Final optimization function:} After applying continuous subset sampling, and approximation of $P(Y|X_{\overline{s}})$ we have simplified our objective to the following equation: 
% \newpage
\begin{equation*}
\min_{\theta } \ E_{X,Y,\zeta }[\log( \mathcal{F}( Z( \theta ,\zeta ) \odot X))]
\end{equation*}
\vspace{-12pt}
\begin{equation}
    =\min_{\theta } \ E_{X,\zeta }\left[\sum ^{c}_{y=1} P( y|X) \log( \mathcal{F}( Z( \theta ,\zeta ) \odot X) )\right]
    \label{eq3}
\end{equation}
$P(y|X)$ is equivalent to $\mathcal{F}(X)$. The expectation operator in the above equation(\ref{eq3}) does not depend on the parameter $\theta$. So, we can learn the parameter $\theta$ by using stochastic gradient descent. We use Adam optimizer in our work. 

\noindent We explain our entire methodology via block diagram in Figure \ref{figbd}. Please note that in the block diagram $X_{\overline{s}}$ refers to $Z( \theta ,\zeta ) \odot X$. 

\vspace{-4pt}
\section{Experiments and Results}
In this section, we quantitatively and qualitatively compare our method with one instance-wise feature selection method L2X and, two pixel attribution methods GradCAM\cite{selvaraju2017grad} and Saliency\cite{simonyan2013deep}. We use MNIST, Fashion MNIST and CIFAR as our datasets for experiments. 

Our method reasons at the level of superpixels or patches of images, which means that our explainer selects patches instead of pixels. Therefore, for the baseline pixel attribution methods we consider the average attribution value for each patch, and then pick top $k$ patches. This is done in order to fairly compare the instance-wise feature selection and existing pixel attribution methods. 

Below we briefly explain the two metrics we use for quantitative evaluation:

\subsection{Post-hoc accuracy\cite{chen2018learning}:}
Each explainability method would return a subset of features/patches $s$ for every instance $x$. Post-hoc accuracy measures how close is the predictive performance of the black-box model when it gets $x_s$ as input w.r.t. getting the entire $x$ as input. It is given by the following formula:
\begin{align*}
& \frac{1}{|\mathcal{X}_{val}|}\sum _{x\in \mathcal{X}_{val}}( \argmax( P( y|x)) ==\argmax( P( y|x_{s}))
\end{align*}
$\mathcal{X}_{val}$ refers to the validation set.

\subsection{Average Causal Effect:}
First, we define individual causal effect(ICE) of a set of features $s$ for a particular instance $x$ as follows:
\begin{equation*}
ICE=P( y|x_{s}) -P( y|x_{random})
\end{equation*}

Here, $x_{random}$ represents an image in which $k$ patches($k=|s|$) belong to $x$ and the rest patches are null. These $k$ patches are selected randomly from $x$. To compute average causal effect(ACE) we simply take the average of the ICE values over the validation set.

\subsection{Results:}
For the all datasets, we report the mean and standard deviation of the post-hoc accuracy and ACE values across 5 runs for L2X and our method. In the tables shown below, $k$ denotes the number of $4\times4$ patches that are selected.
\vspace{2pt}

\textbf{MNIST:}This data set has 28$\times$28 images of handwritten
digits. We use a subset of the classes i.e. class 3 and 8 for experimentation. We train a simple convolutional neural net consisting of 2 layers of convolution and 1 fully connected layer, and achieve 99.74\% accuracy on the test data. We parameterize the selector network of L2X and our method by a 3 layer fully convolutional net. For the L2X, we parameterize its variational approximator by the same network as that of the black-box model(for all the experiments).

\begin{table}[!h]
    \small
    \centering
    \begin{tabular}{|l|l|l|l|}
    \hline
        \textit{Method} & \textit{k=4} & \textit{k=6} & \textit{k=8} \\ \hline
        Our & \textbf{0.953$\pm$ 0.006} & \textbf{0.976 $\pm $ 0.004} & \textbf{0.985$\pm$  0.005} \\ \hline
        L2X & 0.942$\pm$ 0.008 & 0.970$\pm$ 0.004 & 0.981$\pm$ 0.003 \\ \hline
        GradCAM & 0.804 & 0.832 & 0.844 \\ \hline
        Saliency & 0.868 & 0.923 & 0.958 \\ \hline
    \end{tabular}
    \vspace{1pt}
    \caption{Post-hoc accuracy(MNIST)}
    \vspace{-12pt}
\end{table}
\begin{table}[ht]
    \small
    \centering
    \begin{tabular}{|l|l|l|l|}
    \hline
        \textit{Method} & \textit{k=4} & \textit{k=6} & \textit{k=8} \\ \hline
        Our & \textbf{0.351$\pm$  0.012} & \textbf{0.358$\pm$  0.009} & \textbf{0.353$\pm$  0.006} \\ \hline
        L2X & 0.318$\pm$  0.003 & 0.341$\pm$  0.012 & 0.343$\pm$  0.004 \\ \hline
        GradCAM & 0.127 & 0.142 & 0.151 \\ \hline
        Saliency & 0.242 & 0.277 & 0.308 \\ \hline
    \end{tabular}
    \vspace{1pt}
    \caption{Average Causal Effect(MNIST)}
\end{table}

\vspace{-12pt}
\begin{figure}[!h]
\centering
\includegraphics[scale=0.4]{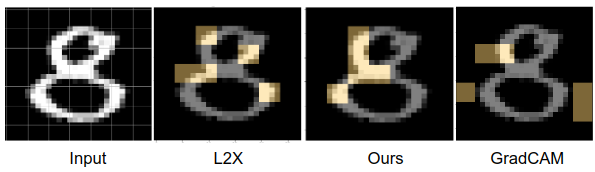}
\includegraphics[scale=0.45]{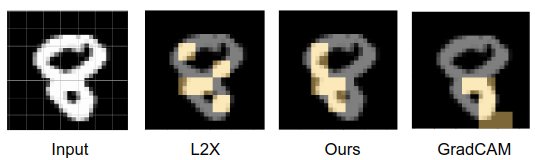}
\caption{In the above figure, we can see that selected patches(highlighted in copper color) of our method capture class discriminative regions(i.e. regions differentiating 3 vs 8) in the query image.(Here, k=5)}
\label{fig2}
\end{figure}
\textbf{FMNIST:}This data set has 28$\times$28 images of fashion items such as t-shirts, shoes, purse etc. We use a subset of the classes i.e. class 0 and 9(t-shirt and shoe) for experimentation. We train a simple convolutional neural net of same architecture as before for MNIST, and achieve 99.9\% accuracy on the test data. We parameterize the selector network of L2X and our method by a 3 layer fully convolutional net. 
\begin{table}[!h]
\small
    \centering
    \begin{tabular}{|l|l|l|l|}
    \hline
        \textit{Method} & \textit{k=4} & \textit{k=6} & \textit{k=8}\\ \hline
        Our & \textbf{0.910$\pm$  0.022} & 0.956$\pm$  0.014 & \textbf{0.978$\pm$  0.005} \\ \hline
        L2X & 0.885$\pm$  0.026 & \textbf{0.963$\pm$  0.006} & 0.970$\pm$  0.013 \\ \hline
        GradCAM & 0.589 & 0.636 & 0.679 \\ \hline
        Saliency & 0.558 & 0.831 & 0.927 \\ \hline
    \end{tabular}
    \vspace{1pt}
    \caption{Post-hoc accuracy(FMNIST)}
\end{table}
\begin{table}[!h]
    \small
    \centering
    \begin{tabular}{|l|l|l|l|}
    \hline
        \textit{Method} & \textit{k=4} & \textit{k=6} & \textit{k=8}\\ \hline
        Our & \textbf{0.177$\pm$  0.009} & \textbf{0.193$\pm$  0.016} & \textbf{0.163$\pm$  0.007} \\ \hline
        L2X & 0.138$\pm$  0.027 & 0.173$\pm$  0.018 & 0.142$\pm$  0.016 \\ \hline
        GradCAM & -0.113 & -0.159 & -0.196 \\ \hline
        Saliency & -0.053 & 0.053 & 0.071 \\ \hline
    \end{tabular}
    \vspace{1pt}
    \caption{Average Causal Effect(FMNIST)}
    \vspace{-5pt}
\end{table}

\begin{figure}[ht]
\centering
\includegraphics[scale=0.25]{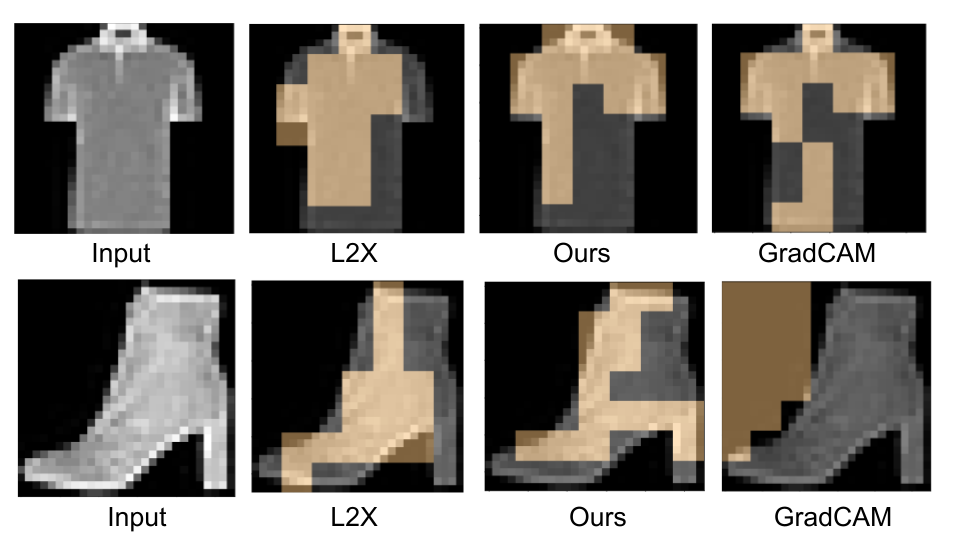}
\caption{Both the rows shows 30\% pixels being selected. Our method appears to focus well on the defining features of the object in the scene(such as heel of the shoe and neck portion of the t-shirt), w.r.t. other methods.}
\label{fig3}
\vspace{-10pt}
\end{figure}

\textbf{CIFAR:}This data set has 32$\times$32 images of 10 different classes. We use a subset of the classes i.e. class 2 and 9(bird and truck) for experimentation. We train a 3 layer convolutional neural net, and achieve 94\% accuracy on the test data. We parameterize the selector network of L2X and our method by a 3 layer fully convolutional net.
\begin{table}[!htp]
    \small
    \centering
    \begin{tabular}{|l|l|l|l|}
    \hline
        \textit{Method} & \textit{20\% pixels} & \textit{30\% pixels} & \textit{40\% pixels} \\ \hline
        Our & \textbf{0.600$\pm$  0.060} & \textbf{0.720$\pm$  0.050} & \textbf{0.780$\pm$  0.030} \\ \hline
        L2X & 0.510$\pm$  0.130 & 0.600$\pm$  0.010 & 0.660$\pm$  0.010 \\ \hline
        GradCAM & 0.580 & 0.660 & 0.710 \\ \hline
        Saliency & 0.551 & 0.570 & 0.610 \\ \hline
    \end{tabular}
     \vspace{1pt}
    \caption{Post-hoc accuracy(CIFAR)}
    \vspace{-10pt}
\end{table}
\begin{table}[!htp]
    \small
    \centering
    \begin{tabular}{|l|l|l|l|}
    \hline
        \textit{Method} & \textit{20\% pixels} & \textit{30\% pixels} & \textit{40\% pixels}\\ \hline
        Our & \textbf{0.078$\pm$  0.055} & \textbf{0.130$\pm$  0.048} & \textbf{0.153$\pm$  0.028} \\ \hline
        L2X & -0.017$\pm$  0.12 & 0.080$\pm$  0.001 & 0.101-0.001 \\ \hline
        GradCAM & 0.055 & 0.08 & 0.088 \\ \hline
        Saliency & 0.028 & 0.033 & -0.005 \\ \hline
    \end{tabular}
     \vspace{1pt}
    \caption{Average Causal Effect(CIFAR)}
    \vspace{-14pt}
\end{table}
\begin{figure}[ht]
\centering
\includegraphics[scale=0.39]{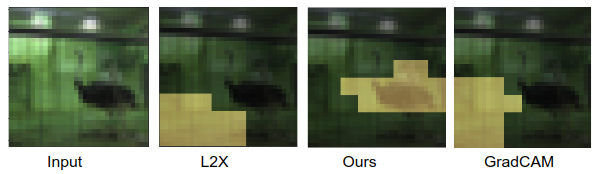}
\includegraphics[scale=0.40]{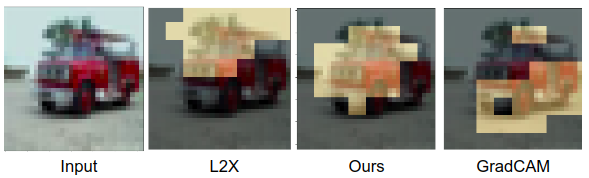}
\caption{The first row shows 20\% pixels being selected, and the second row shows result of selecting 30\% pixels(CIFAR dataset). Our method appears to focus well on the object in the scene, w.r.t. other methods.}
\label{fig3}
\vspace{-10pt}
\end{figure}

\newpage

\section{Related Work}
Currently, in academia and industry\cite{bhatt2020explainable}, feature importance methods are the most prevalent form of explanation methods. Existing methods in this space can be broadly divided into the following categories based on the different algorithmic perspectives taken by them:

(i) \emph{Gradient based methods:} These include methods like Saliency\cite{simonyan2013deep}, Grad-CAM\cite{selvaraju2017grad}, IG\cite{sundararajan2017axiomatic}, SmoothGrad\cite{smilkov2017smoothgrad}. They use gradient information from the trained model to calculate feature attributions. These methods have been widely used in the past, but recent evaluation methods\cite{Adebayo2018SanityCF} show that some of these methods are in-sensitive to the trained model and the training data. Moreover, gradient based methods require computation of gradients and this is not secure in a federated learning setting as simple gradients can be used to generate the original training data\cite{geiping2020inverting}.

(ii) \emph{Shapley-value-based methods:} Several methods exist in this category\cite{vstrumbelj2014explaining,lundberg2017unified,chen2018shapley} the most popular one being SHAP\cite{lundberg2017unified}. SHAP computes the contribution of each feature in an instance to the prediction of the model by using shapley values from coalition game theory. More specifically, it uses locally linear assumption for computing the value function in the shapley value framework. Although this method has been widely adopted for explainability, a recent work\cite{kumar2020problems} shows that using them for feature importance leads to mathematical issues and remedy for that could be causal reasoning.

(iii) \emph{Causal influence based methods:} This is a recent line of work in which causal influence is quantified using some measure and this is used for generating feature attribution maps\cite{chattopadhyay2019neural,schwab2019cxplain,schwab2019granger,o2020generative}. The authors of \cite{chattopadhyay2019neural} interpret a neural network as a SCM and perform causal analysis on the same to find an estimate of the causal strength(via average causal effect) between input and output variables using do-calculus.

(iv) \emph{Instance-wise feature selection methods} This is another recent line of work where the aim is to select features which are most informative w.r.t. the prediction made by the model on a single instance\cite{chen2018learning,yoon2018invase}. 

Our work falls in the intersection of the category of (iii) and (iv) wherein we select most causal features for explaining a black-box model's prediction. Our method only uses the output probabilities of the black-box classifier and thus it is inherently not vulnerable to privacy issues such as those associated with gradient based methods.

\section{Conclusion}
In this work, we derive a causal objective from a rigorously chosen causal strength measure for the task of instance-wise feature subset selection. We also describe a training procedure for solving our causal objective function for real-world experiments. Finally, through carefully chosen metrics we evaluate the proposed method on multiple vision datasets and show its efficacy w.r.t other existing methods. When sparse explanations are required, our method often finds discriminative salient objects. We also provide the code \footnote{\href{https://github.com/pranoy-panda/Causal-Feature-Subset-Selection}{https://github.com/pranoy-panda/Causal-Feature-Subset-Selection}} for reproducing our results. 

{\small
\bibliographystyle{ieee_fullname}
\bibliography{bibtex_file}
}

\end{document}